\documentclass[11pt]{article}

\usepackage[final]{latex/acl}
\usepackage{enumitem}
\usepackage{tabularx}

\usepackage{times}
\usepackage{latexsym}
\usepackage{algpseudocode}
\usepackage{algorithm}

\usepackage[T1]{fontenc}

\usepackage{listings}
\usepackage[utf8]{inputenc}

\usepackage{microtype}

\usepackage{inconsolata}
\usepackage{braket}

\usepackage{graphicx}
\usepackage{booktabs}
\usepackage{multirow}
\usepackage{subcaption}
\usepackage{amsmath} 
\usepackage{amssymb}
\usepackage{bm}
\usepackage{amsmath, amssymb, amsthm}
\usepackage{mathtools}

\usepackage{arydshln}
\newcommand{\mycdashline}[1]{%
    \noalign{\vskip 2pt}  
    \cdashline{#1}        
    \noalign{\vskip 2pt}  
}

\newcommand{\nl}{\mathrm{NL}}
\newcommand{\fl}{\mathrm{FL}}

\title{FormalScience: Scalable Human-in-the-Loop Autoformalisation of Science with Agentic Code Generation in Lean}

\author{
 \textbf{Jordan Meadows\textsuperscript{1}}\;\;\;
 \textbf{Lan Zhang\textsuperscript{2}}\;\;\;
 \textbf{Andr\'e Freitas\textsuperscript{2,3,4}}
\\
 \textsuperscript{1}Independent Researcher
 \\
 \textsuperscript{2}University of Manchester, UK
 \\
 \textsuperscript{3}Idiap Research Institute, Switzerland
 \\
 \textsuperscript{4}National Biomarker Centre, CRUK-MI, UK
\\
 \small{
   \textbf{Correspondence:} \href{mailto:j.c.meadows@hotmail.com}{j.c.meadows@hotmail.com}
 }
}

\begin{document}
\maketitle
\begin{abstract}
Formalising informal mathematical reasoning into formally verifiable code is a significant challenge for large language models. In scientific fields such as physics, domain-specific machinery (\textit{e.g.} Dirac notation, vector calculus) imposes additional formalisation challenges that modern LLMs and agentic approaches have yet to tackle. To aid autoformalisation in scientific domains, we present FormalScience; a domain-agnostic human-in-the-loop agentic pipeline that enables a single domain expert (without deep formal language experience) to produce \textit{syntactically correct} and \textit{semantically aligned} formal proofs of informal reasoning for low economic cost. Applying FormalScience to physics, we construct FormalPhysics, a dataset of 200 university-level (LaTeX) physics problems and solutions (primarily quantum mechanics and electromagnetism), along with their Lean4 formal representations. Compared to existing formal math benchmarks, FormalPhysics achieves perfect formal validity and exhibits greater statement complexity. We evaluate open-source models and proprietary systems on a statement autoformalisation task on our dataset via zero-shot prompting, self-refinement with error feedback, and a novel multi-stage agentic approach, and explore autoformalisation limitations in modern LLM-based approaches. We provide the first systematic characterisation of semantic drift in physics autoformalisation in terms of concepts such as notational collapse and abstraction elevation which reveals what formal language verifies when full semantic preservation is unattainable. We release the codebase together with an interactive UI-based FormalScience system which facilitates autoformalisation and theorem proving in scientific domains beyond physics.\footnote{\url{https://github.com/jmeadows17/formal-science}}

\end{abstract}

\section{Introduction}

The informal mathematical reasoning produced by human researchers deviates significantly from the reasoning expressed in formal languages (FL) compiled by automated theorem provers and formal systems~\cite{iancu2011formalising,mccarthy2022artificial,aliyev2024set}. Formal systems provide strict validation of mathematical statements and proofs at the cost of requiring strict adherence to the syntax of the type-theoretic FL and mathematical library of the prover. This particularly creates barriers towards their broader formalisation in the quantitative sciences~\cite{kaliszyk2015formalizing,meadows2021similarity,bobbin2023formalizingchemicalphysicsusing}. 

The emergence of large language models (LLMs) has partially provided the substrate within which the gap between informal and formal reasoning may be traversed (\textit{i.e.} autoformalisation~\citep{wu2022autoformalization}). However, LLMs have shown increasing hallucination rates with greater task complexity~\cite{opedal2024mathgap,li2025msc} or under out-of-distribution shifts~\cite{stolfo2022causal,meadows2025controlling}, which leads a semantic and syntactic gap when transferring informal representations to formal representations~\cite{zhang-etal-2024-consistent,ganguly2025grammars}. As the problem space deviates further from the mathematics most compatible with the FL, the harder it is to build robust LLM-based autoformalisation methods~\cite{zhang-etal-2025-masa}. These critical limitations inhibit the automation of scientific verification, exploration and fact-checking ~\cite{lu2024ai,yamada2025ai}, making autoformalisation especially challenging in natural sciences such as \textit{Physics}, where semantic drift is significantly magnified by solver incompatibilities with domain notation and calculus.

\begin{figure}[!t]
    \centering
    \includegraphics[width=1\linewidth]{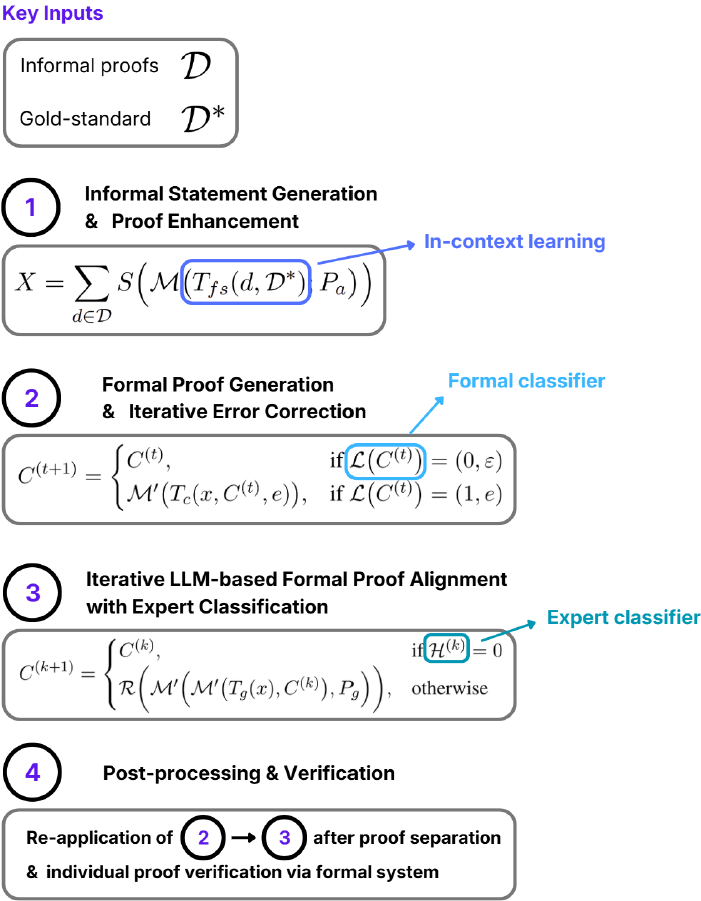}
    \caption{An overview of the \texttt{FormalScience} approach.} 
    \label{fig:overview}
\end{figure}

To understand the bottlenecks of autoformalisation within scientific domains, and using physics as a case study, we:

\noindent \textbf{(i)} Introduce \texttt{FormalScience} (Figure~\ref{fig:overview}): a lightweight, efficient, and cost-effective human-in-the-loop pipeline for scientific statement-proof generation and semi-autoformalisation, which can convert informal reasoning into largely aligned Lean4 code with a 100\% formal guarantee. The pipeline is domain-agnostic and scalable, enabling the generation of high-quality formal datasets in scientific domains for fine-tuning or evaluating AI systems. In this work, we evaluate FormalScience exclusively on physics.


\noindent \textbf{(ii)} Present \texttt{FormalPhysics}: a corpus generated from the FormalScience pipeline comprising of  200 university-level physics statements, informal LaTeX proofs, and formal (Lean4) code. Evaluation using LLM judges confirms that FormalPhysics \textit{matches or exceeds} canonical datasets in formal validity, formal quality, and complexity.


\noindent \textbf{(iii)} Evaluate open-source (up to 30B parameters) and proprietary models on performing autoformalisation of statements in FormalPhysics with three increasingly complex LLM-based approaches: zero-shot autoformalisation, self-refinement with error feedback, and a novel neuro-symbolic multi-stage agentic approach. The leading methodology (GPT-OSS-20B agent) achieves a formal validity of $31\%$ and competitive alignment scores in statement autoformalisation. All other approaches are subject to a distinct alignment-validity trade-off.


\noindent \textbf{(iv)} Provide a detailed qualitative and quantitative analysis of alignment drift in FormalPhysics by characterising drift categories, and using each characterisation to quantitatively describe what Lean4 actually verifies (per physics subdomain) when alignment drift exists yet the formal code successfully compiles, and what formalisation guarantees arise as a consequence. This serves as a step towards explainable verification in scientific autoformalisation.


\section{FormalScience: A human-in-the-loop agentic semi-autoformalisation pipeline}
\label{sec:formalscience}

Autoformalisation refers to the automatic conversion of an informal content (\textit{e.g.} LaTeX) into formally verifiable code, that can be compiled by a solver or formal system (\textit{e.g.} Lean). However, producing syntactically correct and semantically aligned formalisations without human involvement remains a significant challenge (as we later demonstrate). We propose FormalScience (Alg.~\ref{alg:autoformalisation}), a domain-agnostic semi-automatic pipeline for generating problem statements, expanded informal solutions, and formally valid code, thereby facilitating the study of pure autoformalisation. Although applicable to any scientific domain with informal mathematical reasoning (\textit{e.g.} biology, chemistry), all empirical results in this paper are restricted to physics.


We assume a collection of informal proofs $\mathcal{D}$ (\textit{e.g.} LaTeX derivations) and a gold-standard set of informal statement and proofs, $\mathcal{D}^*=[(\mathcal{S}_1, \mathcal{P}_1), ..., (\mathcal{S}_{N'}, \mathcal{P}_{N'})]$ (\textit{e.g.} $N' = 5$). Using in-context learning, a \textit{few-shot template} ($T_{fs}$) is formulated (see Appendix~\ref{app:prompts}) for the purpose of \textbf{(1)} generating statements which correspond to the informal proofs in $\mathcal{D}$; and \textbf{(2)} altering the proofs in $\mathcal{D}$ to align with the gold-standard $\mathcal{D}^*$. This may for instance include the addition of fine-grained derivation steps or textual context. We can write the resulting intermediate dataset as array concatenation 
\begin{equation}
\label{eq:qa_dataset}
    X = \sum_{d \in \mathcal{D}}S\Big(\mathcal{M}\big(T_{fs}(d, \mathcal{D}^*);P_a\big)\Big)
\end{equation}

\noindent where $d$ is a batch of informal proofs, $\mathcal{M}$ is a multi-turn LLM prompting session, and $S$ is a post-processing function splitting the LLM output $\mathcal{M}\big(T_{fs}(d, \mathcal{D^*})\big)$ into separate informal statement-proof pairs, such that $x = [(S_1, P_1), ..., (S_{B'}, P_{B'})]$ (where $B'$ is the batch size, see Alg~\ref{alg:autoformalisation}),  and $x \in X$ aligns with gold-standard $\mathcal{D}^*$. The fixed prompt $P_a$ is used to assess alignment in the prompting session, where the expert suggests improvements and verifies alignment before the next stage.

Next, to obtain the formal code $C$ for each statement-proof pair, we rely on Lean~4 and its Mathlib toolchain to iteratively attempt compilation and return any fatal error messages. We define the tool $\mathcal{L}$ (\textit{i.e.} Lean compiler) such that:
\begin{align}
\label{eq:compile_lean}
    \mathcal{L}(C) = \begin{cases}
            (0, \varepsilon), \text{ if $C$ compiles}\\
            (1, \text{e}), \text{ otherwise}
           \end{cases}
\end{align}
where $e$ is the error message for a given iteration, and $\varepsilon$ is the empty string. To begin iterative error correction, we initially use a code generation template $T_{g}$ to prompt an LLM-based agent $\mathcal{M'}$ to use tool $\mathcal{L}$, with correction template $T_{c}$. If we define the first output as $C^{(0)} = \mathcal{M}'\big(T_{g}(x)\big)$, we obtain compilable code via
\begin{equation}
\resizebox{1\columnwidth}{!}{$\displaystyle
    C^{(t+1)} = \begin{cases}
        C^{(t)}, &\text{if } \mathcal{L}\big(C^{(t)}\big) = (0, \varepsilon) \\
        \mathcal{M}'\big(T_{c}(x, C^{(t)}, e)\big), &\text{if } \mathcal{L}\big(C^{(t)}\big) = (1, e)
    \end{cases}
$}
\label{eq:correction_iteration}
\end{equation}
\noindent which terminates when $t^* = \min \{t:\mathcal{L}\big(C^{(t)}\big) = (0, \varepsilon)\}$, to give $C = C^{(t^*)}$. In practice, the multi-turn conversational agent $\mathcal{M}'$ operates over a representation of the chat history contained within its context window. We can write this correction loop in shorthand as $C = \mathcal{R}\big(C^{(0)}\big)$.

The previous phase does not consider whether semantic drift occurs between the informal and formal representations. Defining the LLM-based evaluation of the formal code's alignment with the initial prompt ($T_g(x)$) as $\mathcal{M}'\big(T_g(x), C\big)$, we can write the \textit{expert alignment classification} step at iteration $k$ as 
\begin{equation}
    \mathcal{H}^{(k)} = \mathcal{H}\Big(\mathcal{M}'\big(T_g(x), C^{(k)}\big)\Big) \in \{0,1\}
\end{equation}
\noindent where the human intervenes as a binary classifier (analogous to compilation tool $\mathcal{L}(C)$). We obtain both \textit{aligned} and \textit{corrected} code via the following iterative process
\begin{equation}
\label{eq:alignment_iteration}
\resizebox{1\columnwidth}{!}{$\displaystyle
    C^{(k+1)} = \begin{cases}
        C^{(k)}, &\text{if } \mathcal{H}^{(k)} = 0\\
        \mathcal{R}\bigg(\mathcal{M}'\Big(\mathcal{M}'\big(T_g(x), C^{(k)}\big), P_g \Big)\bigg), &\text{otherwise}
    \end{cases}
$}
\end{equation}
which terminates when $k^* = \min \{k:\mathcal{H}^{(k)} = 0\} \in [0, \mathcal{P}]$, where patience $\mathcal{P}$ denotes the maximum number of expert alignment classification iterations, the fixed prompts $P_a$ and $P_g$ are respectively used provide an LLM-based assessment of alignment and generate improvements, and the code is iteratively corrected via $\mathcal{R}$ (Eq.~\ref{eq:correction_iteration}). 

Finally, a post-processing step is used to extract individual proofs from each output generated by Eq.~5, which results in tuples $(S, P, C)$. However, the post-processing may have introduced errors, so all $C$ are recompiled with $\mathcal{L}$ to find invalid formal proofs, which are iteratively improved via Eq.~3-5.

\section{FormalPhysics: Physics Formalization in Lean4}

We select physics as the target scientific domain and Lean as the target formal language, and apply FormalScience using GPT-5.1 and Claude-Opus-4.5~\cite{anthropic2025claudeopus45}. The resulting dataset (FormalPhysics) is a corpus containing 200 physics statements, informal LaTeX proofs, and complete formal proofs. This scale is consistent with established autoformalisation test sets such as miniF2F~\citep{zheng2022miniff} (244 test examples) and ProofNet~\citep{azerbayev2023proofnetautoformalizingformallyproving} (371 examples), while containing approximately twice as many mathematical objects and formulae per example (Table~\ref{tab:data_stat}). FormalPhysics is intended as an evaluation benchmark rather than a fine-tuning corpus. The examples used as input to FormalScience to generate FormalPhysics are sourced from related work~\cite{meadows-etal-2024-exploring}, and the human-in-the-loop pipeline was conducted by one physics expert within one month at a total cost of approximately 50 USD. With a motivated group of experts using FormalScience, they could generate thousands of verified \textit{field-specific} formalisations from scientific works in a similar timeframe. Such larger-scale data could be used to fine-tune custom LLMs and improve or evaluate the capabilities of state-of-the-art mathematical discovery approaches. The exact implementation of Alg.~\ref{alg:autoformalisation} used to construct FormalPhysics is described in Appendix~\ref{app:autoformalisation-pipeline-details}.

\subsection{Benchmark Comparison}

\begin{table*}[h!]
    \small
    \centering
    \begin{tabular}{l c c c c c c}
        \toprule
        \textbf{Dataset} & \textbf{Size} & \textbf{Domain} & $\bf{s_\nl}$ & $\bf{p_\nl}$ & $\bf{s_\fl}$ & $\bf{p_\fl}$\\
        \midrule
        miniF2F~\citep{zheng2022miniff} & 488 & Olympiad (Ol) Math & \checkmark & \checkmark & \checkmark & Partial\\
        ProofNet~\citep{azerbayev2023proofnetautoformalizingformallyproving} & 371 &  Undergraduate (UG) Math & \checkmark & \checkmark & \checkmark & $\times$\\
        Lean-Dojo~\citep{yang2024leandojo} & 98,734 & Mathlib & $\times$ & $\times$ & \checkmark  & \checkmark\\
        Lean Workbook~\citep{ying2024lean} & 57,231 & High-School Math & \checkmark & $\times$ & \checkmark & Partial\\
        FormalMATH~\citep{yu2025formalmathbenchmarkingformalmathematical} & 5,560 & Ol \& UG Math & \checkmark & \checkmark & \checkmark & $\times$\\
        Herald-Statement~\citep{gao2025herald} & 579,883 & Mathlib & \checkmark & $\times$ & \checkmark & $\times$\\
        Herald-Proof~\citep{gao2025herald} & 44,553 & Mathlib & \checkmark & \checkmark & \checkmark & \checkmark\\
        \midrule
        FormalPhysics & 200 & Advanced Physics & \checkmark & \checkmark & \checkmark & \checkmark\\
        \bottomrule
    \end{tabular}
    \caption{Properties of Lean4 formal benchmarks. $s_\nl$: Natural Language Statement; $p_\nl$: Natural Language Proof; $s_\fl$: Formal Language Statement; $p_\fl$: Formal Language Proof.}
    \label{tab:data_property}
\end{table*}

\begin{table*}[h!]
    \small
    \centering
    \begin{tabular}{l c c c c c c}
        \toprule
        & \multicolumn{2}{c}{$s_\nl$ Complexity} & \multicolumn{2}{c}{$s_\fl$ Correctness} & \multicolumn{2}{c}{$s_\nl$-$s_\fl$ Alignment}\\
        \mycdashline{2-7}
        \textbf{Dataset} & \textbf{Objects} & \textbf{Formulae} & \textbf{FV} (\%) & \textbf{FQ} (\%) & \textbf{LP} (\%) & \textbf{MC} (\%)\\
        \midrule
        miniF2F & 3.14$\pm$1.55 & 3.21$\pm$1.53 & 88.00 & 63.00 & 92.00 & 92.00\\
        ProofNet & 3.67$\pm$1.48 & 3.62$\pm$1.52 & 95.50 & 61.50 & 77.50 & 77.50\\
        Lean Workbook & 3.67$\pm$1.99 & 3.62$\pm$2.26 & 89.00 & 46.00 & 78.00 & 85.00\\
        FormalMATH & 4.47$\pm$2.45 & 4.53$\pm$2.62 & 97.50 & \textbf{80.00} & \textbf{98.00} & \textbf{96.50}\\
        Herald-Statement & 4.92$\pm$2.43 & 4.80$\pm$2.30 & 80.50 & 63.50 & 87.00 & 87.00\\
        Herald-Proof & \textbf{6.57}$\pm$2.32 & \textbf{6.42}$\pm$2.37 & 2.00 & 73.00 & 94.50 & 94.00\\
        \midrule
        FormalPhysics & 6.41$\pm$2.34 & 6.22$\pm$2.13 & \textbf{100.00} & 73.50 & 72.00 & 72.50\\
        \bottomrule
    \end{tabular}
    \caption{Statistics derived from 200 examples randomly selected from each dataset. \textbf{Objects}: How many math or physics objects excluding explicit numbers and variables are mentioned directly in the natural language statement? \textbf{Formulae}: How many math or physics formulae are mentioned directly in the natural language statement? \textbf{FV}: Formal Validity (\textit{i.e.} Pass rate); \textbf{FQ}: Formal Quality; \textbf{LP}: Logical Preservation; \textbf{MC}: Mathematical Consistency.}
    \label{tab:data_stat}
\end{table*}

We compare the properties of our dataset with existing benchmarks for formal mathematics. All datasets in Table~\ref{tab:data_property} contain a formal language statement ($s_\fl$) (\textit{e.g.} Lean4 statement), which is the formal representation of a given natural language statement ($s_\nl$). \textit{Statement Autoformalisation} is the task of automatically translating the NL statement to the FL statement ($s_\nl \rightarrow s_\fl$), yet only 7/8 datasets contain an NL statement. A natural language proof ($p_{\nl}$) is an informal proof of a given NL statement. Automating this reasoning ($s_{\nl} \rightarrow p_{\nl}$) is typical of \textit{scientific/mathematical QA tasks}, yet only 5/8 datasets contain an informal NL proof (\textit{e.g.} LaTeX derivation). Only 3/8 datasets contain full FL proofs ($p_\fl$) (\textit{e.g.} without "sorries"), thereby supporting \textit{automated theorem proving} and \textit{full theorem autoformalisation} tasks. Only FormalPhysics and Herald-Proof are compatible with every task. 

We further randomly sample 200 examples from each benchmark and evaluate them across different dimensions, and use elements from a taxonomy for autoformalisation evaluation~\cite{zhang2025goldstandardsepistemicensemble} to evaluate each sample. Through this dual approach, both the \textit{syntactic quality} of the generated code and its \textit{semantic alignment} with the input natural language (NL) statement are evaluated.

\paragraph{Evaluation methodology.} For a given input example, each element can be measured via a \textit{binary classification} with respect to a target characteristic. Classification results are averaged over the sample to give the percentages in Table~\ref{tab:data_stat}. \textbf{Formal Validity} (FV) is judged by the Lean4 theorem prover, while the remaining metrics are assessed via LLM-as-a-judge. \textbf{Formal Quality} (FQ) is the rate that the formal code is of high quality in regard to structural clarity and usefulness. \textbf{Logical Preservation} (LP) is the rate that the code captures the logical structure and content of the original NL statement. \textbf{Mathematical Consistency} (MC) is the rate that the formal code accurately represents mathematical objects and operations present in the NL statement. We also measure the number of objects and formulae in the natural lanugage statements with LLM judges. We prompt GPT-4.1-mini with 0.2 temperature to obtain judgments. Although there might be some noise in the judgments, the underlying LLM is still unbiased towards a specific benchmark and results are still indicative. To assess the robustness of these judgments, we conduct an inter-judge agreement analysis using an independent second judge (Qwen2.5-Coder-7B-Instruct) across ${\sim}6{,}000$ paired binary judgments (Appendix~\ref{app:judge-robustness}). Both judges unanimously rank GPT-5.1 first on every metric in every setting, and the alignment-validity trade-off holds under both judges ($\rho \in [-0.10, 0.30]$, all $p > 0.6$). The 7B judge is systematically more conservative in absolute scores but detects the same underlying quality signal (phi coefficients $0.28$--$0.37$, all $p < 10^{-19}$).



\paragraph{Formalisation in physics is more complex.} A given natural language statement in FormalPhysics contains on average \textit{twice as many mathematical objects and formulae} as miniF2F~\cite{zheng2022miniff}, ProofNet~\cite{azerbayev2023llemma}, and Lean Workbook~\cite{ying2024lean}. FormalPhysics also contains $\approx 33\%$ more objects/formulae than the recent FormalMath~\cite{yu2025formalmathbenchmarkingformalmathematical} and Herald-Statement~\cite{gao2025herald} datasets per example, matched only by Herald-Proof (Table~\ref{tab:data_stat}).

\paragraph{High formal validity with examples generated through FormalScience.} All formal code examples generated from our pipeline are syntactically valid on the latest Lean4 version (see FV, Table~\ref{tab:data_stat}). Notably, miniF2F and ProofNet were released circa 2023 yet respectively score $88\%$ and $96\%$ successful compilation rates (on an older version of Lean). Herald-Proof~\cite{gao2025herald} achieves only $2\%$ FV, while FormalMATH~\cite{yu2025formalmathbenchmarkingformalmathematical} scores $98\%$. FormalPhysics obtains a perfect score.

\paragraph{High formal quality (FQ), low logical preservation (LP) and mathematical consistency (MC).} FormalPhysics obtains the second highest score for formal quality (a reference-free evaluation of formal code quality), while scoring the lowest for both logical preservation and mathematical consistency with the NL statement. This is a natural consequence stemming from the fact that around half of NL statements within FormalPhysics contain vector calculus (\textit{i.e.} electromagnetism) or Dirac notation (\textit{i.e.} quantum mechanics) with non-commutative operators. Lean4 does not directly support vector calculus or Dirac notation, and struggles with basic derivatives and integrals~\cite{bobbin2023formalizingchemicalphysicsusing}, hence alternative strategies and notation were required to successfully compile the code. Formalising physics is extremely challenging~\cite{kaliszyk2015formalizing} and we qualitatively explore such misalignment in Section 5.

\section{Experimental Results}
Towards fully automated and formally verifiable scientific reasoning, we use FormalPhysics for statement autoformalisation with both open-source and proprietary LLMs in three increasingly complex inference pipelines: \textbf{(1)} zero-shot prompting; \textbf{(2)} self-refinement with error feedback; \textbf{(3)} an agentic code generation approach based on a recent framework~\cite{wang2024executable}. The results are provided in Table~\ref{tab:llm-baselines}.

\paragraph{Models.} We use various open-source models such as Qwen2.5-Coder-7B~\citep{hui2024qwen25codertechnicalreport}, DeepSeek-Prover-V2-7B~\citep{ren2025deepseekproverv2advancingformalmathematical}, Kimina-Autoformalizer-7B~\citep{wang2025kiminaproverpreviewlargeformal}, GPT-OSS-20B~\citep{openai2025gptoss120bgptoss20bmodel}, a distillation of Claude-Sonnet-4.5~\cite{anthropic2025claudesonnet45} onto Qwen3-14B~\cite{teichai2025qwen3_sonnet45, yang2025qwen3technicalreport}, Qwen3-Coder-30B, in addition to the proprietary GPT-5.1~\citep{openai_gpt51_2025}. 

\begin{table}[t!]
  \centering
  \resizebox{1\columnwidth}{!}{%
  \begin{tabular}{l c c c c}
    \toprule
    \textbf{LLM} & \textbf{FV} (\%)& \textbf{FQ} (\%)& \textbf{LP} (\%)& \textbf{MC} (\%)\\
    \midrule
    \multicolumn{5}{l}{\textbf{(1)} \;\; \texttt{Zero-Shot Autoformalisation}}\\
    \midrule
    Qwen2.5-Coder-7B & 1.00 & 15.00 & 24.00 & 20.50\\ 
    DeepSeek-Prover-7B & 13.00 & 23.00 & 27.50 & 24.00\\ 
    Kimina-7B & \textbf{51.50} & 6.50 & 10.50 & 9.50\\
    GPT-OSS-20B & 4.50 & 68.50 & 73.00 & 72.50\\
    GPT-5.1 & 14.50 & \textbf{79.50} & \textbf{76.50} & \textbf{77.00}\\
    \midrule
    \multicolumn{5}{l}{\textbf{(2)} \;\; \texttt{Self-Refinement with Error Feedback}}\\
    \midrule
    Qwen2.5-Coder-7B & 1.00 & 16.50 & 23.00 & 19.50\\
    DeepSeek-Prover-7B & 4.50 & 17.00 & 23.00 & 23.00\\
    Kimina-7B & \textbf{23.00} & 6.50 & 9.50 & 8.00\\
    GPT-OSS-20B & 7.50 & 70.50 & 77.00 & 79.00\\
    GPT-5.1 & 17.00 & \textbf{82.50} & \textbf{82.00} & \textbf{82.00}\\
    \midrule
    \multicolumn{5}{l}{\textbf{(3)} \;\; \texttt{Agentic Code Generation Pipeline}}\\
    \midrule
    Qwen3-Sonnet-14B & \textbf{52.00} & 1.00 & 10.50 & 6.50\\
    GPT-OSS-20B & 31.00 & \textbf{73.00} & \textbf{72.50} & \textbf{73.00}\\
    Qwen3-Coder-30B & 5.50 & 49.50 & 59.00 & 48.00\\
    \midrule
    \multicolumn{5}{l}{\textbf{(4)} \;\; \texttt{FormalScience} (ours)}\\
    \midrule
    GPT-5.1 / Claude-4.5 & \textbf{100.00} & 73.50 & 72.00 & 72.50\\
    \bottomrule
  \end{tabular}
  }
  \caption{Performance of LLM-based approaches on the FormalPhysics corpus (using GPT-4.1-mini).}
  \label{tab:llm-baselines}
\end{table}

\begin{table}[t!]
  \centering
  \resizebox{1\columnwidth}{!}{%
  \begin{tabular}{l c c c c}
    \toprule
    \textbf{LLM} & \textbf{FV} (\%)& \textbf{FQ} (\%)& \textbf{LP} (\%)& \textbf{MC} (\%)\\
    \midrule
    \multicolumn{5}{l}{\textbf{(1)} \;\; \texttt{Zero-Shot Autoformalisation}}\\
    \midrule
    Qwen2.5-Coder-7B & 1.00 & 8.00 & 9.00 & 12.50\\ 
    DeepSeek-Prover-7B & 13.00 & 12.50 & 13.50 & 14.00\\ 
    Kimina-7B & \textbf{51.50} & 11.00 & 14.50 & 6.50\\
    GPT-OSS-20B & 4.50 & 15.50 & 12.50 & 17.50\\
    GPT-5.1 & 14.50 & \textbf{27.00} & \textbf{28.00} & \textbf{33.00}\\
    \midrule
    \multicolumn{5}{l}{\textbf{(2)} \;\; \texttt{Self-Refinement with Error Feedback}}\\
    \midrule
    Qwen2.5-Coder-7B & 1.00 & 11.50 & 7.00 & 10.50\\
    DeepSeek-Prover-7B & 4.50 & 26.50 & 11.50 & 17.00\\
    Kimina-7B & \textbf{23.00} & 6.00 & 7.50 & 5.00\\
    GPT-OSS-20B & 7.50 & 14.50 & 10.50 & 16.50\\
    GPT-5.1 & 17.00 & \textbf{38.00} & \textbf{35.00} & \textbf{42.00}\\
    \bottomrule
  \end{tabular}
  }
  \caption{Performance of LLM-based approaches on the FormalPhysics corpus (using Qwen2.5-Coder-7B-Instruct).}
  \label{tab:llm-baselines-7b}
\end{table}

\subsection{Zero-shot and self-refinement pass}

We discuss physics autoformalisation baselines for settings \textbf{(1)} and \textbf{(2)} (prompts in Appendix~\ref{app:prompts}). The self-refinement baselines are based on the zero-shot formalisations from the same LLM.


\paragraph{Trade-off between formal validity and semantic alignment.} The Spearman and Pearson coefficients of FV and the mean of FQ, LP, and MC are both zero to one decimal place (with $p > 0.9$), indicating that an approach with high probability of generating syntactic valid formalisations for physics will struggle to simultaneously represent the intended semantics of the problem. 

\paragraph{Invariance to naive prompting.} The LLM-as-a-judge scores (FQ, LP, MC) are effectively unchanged per model between the zero-shot and self-refinement settings under the primary GPT-4.1-mini judge. The error-based self-refinement method uses error details to improve zero-shot output, at the cost of $\approx 2$x the token usage, without clear improvement to formal validity or alignment scores for FormalPhysics. However, this invariance is judge-dependent: under an independent 7B judge (Appendix~\ref{app:judge-robustness}), GPT-5.1 gains +9.0pp and Kimina-7B drops $-$4.5pp between settings.

\subsection{Agentic code generation}
To establish a best-effort open-source baseline on consumer-grade hardware, we implement an agentic code generation pipeline system aiming to maximise compilation rates and alignment without human intervention. We provide a full derivation of our implementation and further details in Appendix~\ref{app:agent-derivation} and Alg.~\ref{alg:agentic}. Each baseline in Table~\ref{tab:llm-baselines} required \textbf{100+ hours of compute on a 5090 RTX GPU} (\textit{i.e.} 30+ minutes per Physics proof). This is approximately equivalent to the code generation rate of the FormalScience approach.

We use an LLM as the base model for a CodeAgent within the smolagents framework~\cite{wang2024executable}. The agent may use Python functions as tools during inference (in a ReAct~\cite{yao2022react} cycle) which generally features a planning step, a tool-calling action step, and an observation step where the model assesses the tool's output. The agent may output a final answer based on the observation or begin another cycle.

Our implementation features two primary stages. First, an initial generation phase outputs Lean code which is fed to a surface guard that rejects code containing forbidden tokens, incomplete proofs, or malformed imports before compilation. Second, an iterative correction phase compiles the code and categorises errors. \textit{Structural} errors (syntax, unknown identifiers, missing modules) trigger full regeneration using the base LLM with hints based on error type, while \textit{semantic} errors (type mismatches, unsolved goals) are addressed using a patch agent that applies minimal unified diffs. It terminates after 25 correction iterations (full ReAct cycles) or successful compilation in Lean.

The 7B models (Kimina-Autoformalizer, DeepSeek-Prover) were excluded from the agentic setting due to insufficient base capability. DeepSeek-Prover-7B actually \textit{decreases} to 4.5\% FV under self-refinement, suggesting it cannot effectively incorporate error feedback even in the simplest iterative setting. Kimina-7B achieves high FV (51.5\%) but the lowest alignment scores (FQ: 6.5\%, LP: 10.5\%), indicating it exploits compilation shortcuts without capturing physics semantics. The multi-step planning, error categorisation, and diff generation required by the agentic pipeline would compound rather than resolve these limitations.

\paragraph{Open-source models can overcome the alignment-validity trade-off.} The previous GPT-OSS-20B baselines can only produce less than 10\% formally valid formalisations. The agentic \textit{approach improved this to $31\%$} without any significant decrease in LLM-as-a-judge scores.

\paragraph{Autoformalisation is an emergent capability dependent upon parameter count, neuro-symbolic integration, and test-time scaling.} FormalScience obtained a formal validity score of over $3$x the best open-source agentic approach. Furthermore, the results are highly sensitive to the base LLM choice, where larger models (\textit{e.g.} 14B, 30B) do not necessarily outperform the naive prompting approaches utilising smaller models (\textit{e.g.} Kimina-7B). \textit{Without iterative dialogue with a symbolic prover}, it is difficult for a leading transformer-based LLM to produce physics formalisations with high formal validity, regardless of the test-time scaling techniques used. Similarly, there exists a minimum base LLM reasoning capability (determined by parameter count, context window, etc.) required to make effective use of symbolic tools (e.g. minimal unified diff) in agentic pipelines. Without any/minimal test-time scaling (\textit{e.g.} zero-shot) GPT-5.1 obtained only $15\%$ formal validity, yet when used within FormalScience it performed significantly better. These three extremes (no symbolic tools, low base LLM intelligence, no test-time scaling) demonstrate the type of experimental optimisation problem required to deliver physics autoformalisation pipelines. The qualitative separation between large and small models persists across judges, though the magnitude of the gap is judge-dependent (Appendix~\ref{app:judge-robustness}).

\section{Alignment and Qualitative Analysis}

\begin{figure*}
    \centering
    \includegraphics[width=\textwidth]{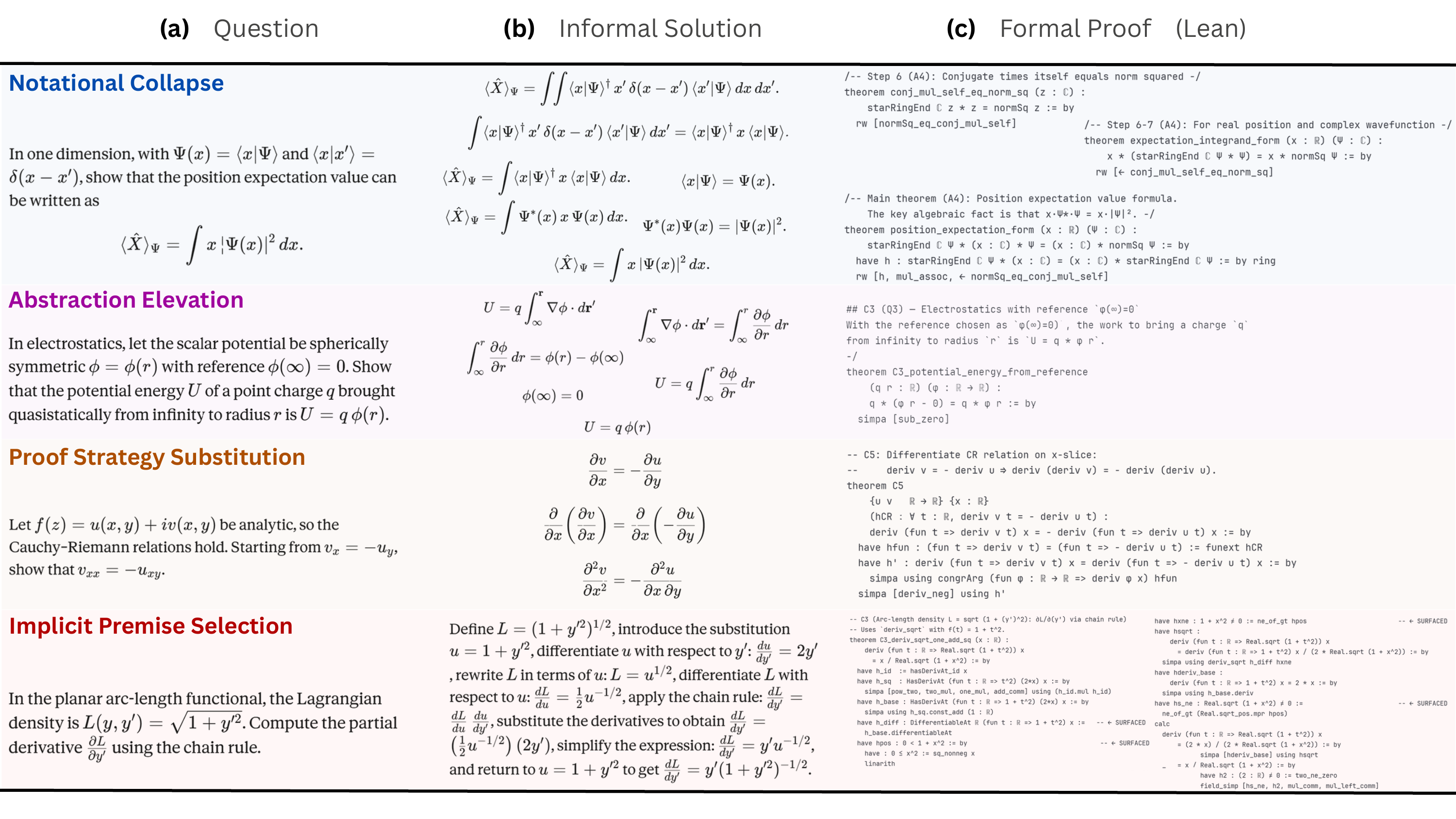}
    \caption{Examples sourced from the FormalPhysics corpus generated via the FormalScience approach applied to Physics. Each row is associated with a different class of semantic drift (\textit{e.g.} Notational Collapse, Abstraction Elevation). See Tab.~\ref{tab:detailed-alignment-table} for a detailed analysis and additional examples.} 
    \label{fig:human_vs_lean}
\end{figure*}

When a syntactically valid formalisation uses fundamentally different mathematical objects, \textit{what exactly has been verified?} What partial guarantees can formalisation provide when full semantic preservation is unattainable? To answer this we quantitatively measure alignment divergence using a distinct categorisation schema, and use this to guide a qualitative characterisation of the semantic drift induced by formalisation to Lean. We define the following drift categories (visualised in Fig.~\ref{fig:human_vs_lean}):\newline

\noindent \textbf{Notational Collapse}: Domain-specific physics notation collapsed to simpler mathematical objects.

The relevant Fig.~\ref{fig:human_vs_lean} example expects a solution integrating over the continuous delta function and multiple substitution operations with terms defined using Dirac/braket notation. The Lean proof correctly defines $x$ as a real scalar but inappropriately defines the quantum state \textit{vector} $\ket{\Psi}$ as the complex scalar $\Psi$. Quantum mechanics is formalised within a complete, complex inner-product (\textit{i.e.} Hilbert) space, which is enforced implicitly through Dirac notation. When $\ket{\Psi}$ is collapsed to $\Psi$ in this manner (\textit{i.e.} ignoring $\braket{x|\Psi} = \Psi(x)$) the fundamental formalism of QM is not respected. 

\textbf{What did Lean verify?} Essentially $z\cdot z^* = |z|^2$ for $z \in \mathbb{C}$. All quantum mechanical formalism (and calculus) is absent. \newline

\noindent \textbf{Abstraction Elevation}: Symbolic operations replaced by abstract algebraic properties. 

The Fig.~\ref{fig:human_vs_lean} example requires the evaluation of a definite line integral in 3-dimensional space given the assumption that the scalar potential $\phi (r) = 0$ as $r \rightarrow \infty$. The Lean proof correctly defines a real function $\phi$ with real scalars $q$ (charge) and $r$ (radius), but skips all vector calculus.

\textbf{What did Lean verify?} That $x - 0 = x$ for $x \in \mathbb{R}$. The logic is that $U = q\phi(r)$ (goal statement), so the potential difference between a charge at $r$ and at $\infty$ is $\Delta U = q\phi(r) - q\phi(\infty) = q\phi(r) - 0 = q\phi(r)$. The physics has been abstracted into the hypothesis statement given in the question. \newline

\noindent \textbf{Proof Strategy Substitution}: Theorem proved via alternative approach to the informal derivation.

The informal solution (Fig.~\ref{fig:human_vs_lean}b) applies a direct partial differentiation operator to a complex function (\textit{i.e.} a single operation). Lean uses an alternative (correct) strategy to show that if two functions are equal their derivatives must be equal.  

\textbf{What did Lean verify?} The target statement $\frac{\partial^2 v}{\partial x^2} = -\frac{\partial^2 u}{\partial x \partial y}$, but circumvented direct differentiation of the supporting premise. \newline

\noindent \textbf{Implicit Premise Selection}: Assumptions that are unstated in the NL statement or derivation, yet are explicitly defined as FL hypotheses.

The question requires the differentiation of a Lagrangian via the chain rule. Lean provides \textit{a deeper proof by surfacing a number of implicit premises} (\textit{e.g.} $1 + y' > 0 \; \forall y \in \mathbb{R}$, do not divide by zero). This reveals the hidden logical structure of the argument often ignored by physicists.

\textbf{What did Lean verify?} Lean effectively verified $\partial_x(\sqrt{1 + x^2}) = x(1 + x^2)^{-1/2}$ \textit{without using the chain rule} by introducing relevant premises.

\begin{figure}[t]
    \centering
    \includegraphics[width=1\linewidth]{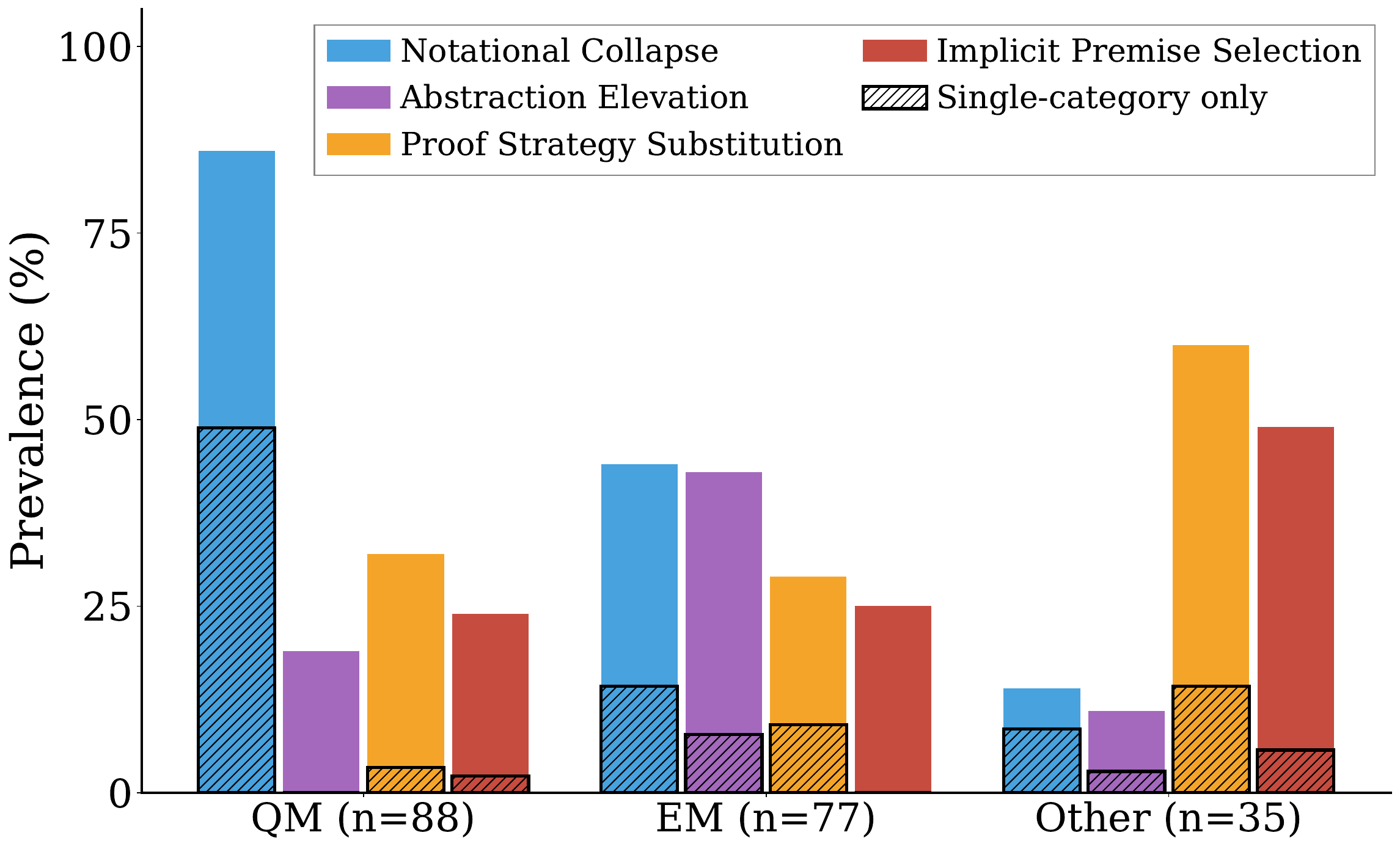}
    \caption{Proportion of alignment drift categories present in FormalPhysics examples by Physics subdomain: Quantum Mechanics (QM), Electromagnetism (EM), and Other (\textit{e.g.} classical, statistical). \textbf{Single-category only}: Prevalence of examples featuring only one type of drift.} 
    \label{fig:alignment_results}
\end{figure}

\paragraph{What is Lean verifying quantitatively?} Fig.~\ref{fig:alignment_results} describes the prevalence of each semantic drift category by Physics subdomain. Notational Collapse is present in $> 75\%$ of all QM proofs. This is the most severe type of drift because fundamental scientific context is ignored due to the mistranslation of semantically dense mathematical objects. These proofs verify a \textit{subset} of the required argumentation using \textit{mathematical objects with simple types and structure}.

Abstraction Elevation occurs in $\approx 25\%$ across FormalPhysics. Meaningful physics calculations are replaced by simple abstract arguments often proving trivial results (\textit{e.g.} $x - 0 = x$). In addition, Tab.~\ref{tab:detailed-alignment-table} describes a pattern where vector calculus identities are replaced by abstract linear maps. When vector calculus is formalised as abstract linear algebra, then compilation \textit{verifies algebraic coherence within an abstract vector space}. This does not respect specifics such as coordinate geometry or \textit{e.g.} verify Maxwell's equations. These proofs \textit{verify that a solution is possible} but not necessarily the goal solution outlined in the question.

Proof Strategy Substitution is present within $\approx 33\%$ of FormalPhysics. These proofs \textit{verify statements where the original mathematical objects are preserved} while the goal conclusion \textit{is} verified using a different strategy.

Implicit Premise Selection is the only unambiguously beneficial alignment drift and occurs in $\approx 25\%$ of examples, where only $2\%$ are ``pure'' (no other drift present). The pure examples are actually of \textit{higher quality} than their informal counterparts.


\section{Related Work}
Recent advances in large language models (LLMs) have renewed interest in autoformalisation, which aims to bridge informal mathematical language and formal proof systems~\citep{wu2022autoformalization,yang2025position,mensfelt2025commonframeworkautoformalization,zhang-etal-2025-autoformalization}. Prior work has demonstrated the feasibility of translating natural-language mathematical statements into formal representations and proofs. Autoformalisation has been applied to a range of tasks, including verification of natural-language explanations in natural language inference~\citep{quan-etal-2024-verification,quan-etal-2024-enhancing} and the construction of automated theorem proving pipelines~\citep{jiang2023draft,tarrach2024more,liu2025bootstrapping}. Recent systems increasingly incorporate retrieval-augmented generation, which improves correctness and consistency by leveraging existing formal libraries, in both Isabelle~\citep{zhang-etal-2024-consistent} and Lean~\citep{yang2023leandojotheoremprovingretrievalaugmented,liu2025rethinking,wang2025improvingautoformalizationusingdirect,zhang2025driftdecomposeretrieveillustrate}. Complementary approaches include process-driven frameworks that structure the formalization pipeline~\citep{lu2024processdrivenautoformalizationlean4} and self-consistency methods for selecting high-quality outputs~\citep{li2024autoformalize}.

On the benchmarking side, existing datasets exhibit a trade-off between quality and scale. Human-curated benchmarks~\citep{zheng2022miniff,azerbayev2023proofnetautoformalizingformallyproving,tsoukalas2024putnambench,poiroux-etal-2025-reliable} offer high-quality annotations but are too small for large-scale training. Benchmarks derived from formal libraries~\citep{yang2024leandojo,zhang-etal-2024-consistent,xin2025apebenchifilelevelautomated} scale better but risk data contamination due to overlap with model pretraining. Automated data generation pipelines~\citep{jiang2024multilanguage,ying2024lean,yu2025formalmathbenchmarkingformalmathematical,gao2025herald,liu2025atlasautoformalizingtheoremslifting} address scalability but often produce shallow or low-quality formalizations. Our work targets this gap by aiming to improve or characterise the quality of autoformalisation data while preserving scalability and minimising contamination.

We note that existing state-of-the-art autoformalisation systems such as DRIFT~\citep{zhang2025driftdecomposeretrieveillustrate} and retrieval-based approaches are evaluated on Mathlib-derived benchmarks where library coverage is comprehensive. FormalPhysics targets domains (vector calculus, Dirac notation, non-commutative operators) where Mathlib support is absent, making direct comparison methodologically inappropriate. Crucially, many related methods do not perform semantic alignment evaluation, which is a central contribution: demonstrating that formal validity alone is insufficient for physics autoformalisation.

\section{Conclusion}

For the purpose of exploring autoformalisation limitations in science, we propose a human-in-the-loop agentic methodology (\texttt{FormalScience}) for formalising scientific reasoning in Lean. Applying it to physics, we produce a dataset (\texttt{FormalPhysics}) comprising 200 questions, informal solutions, and formal proofs across quantum mechanics, electromagnetism, and other subdomains. We use compilation success rates and LLM-as-a-judge metrics to compare the formal quality and alignment of formalised statements across several notable autoformalisation benchmarks, and find FormalPhysics leads with $100\%$ formal verification rate (and is competitive in other metrics) at the cost of alignment drift. We argue alignment issues are due to the incompatibility of formal systems with domain-specific machinery such as vector calculus and Dirac notation. 

We use FormalPhysics to test the Physics autoformalisation and question-answering capabilities of three increasingly complex LLM-based inference pipelines, including zero-shot, error-driven self-refinement, and full agentic code generation involving notation surface guards, prompt regeneration based on categorised Lean compilation errors, and iterative self-refinement via a patch agent utilising the ReAct~\cite{yao2022react} framework. We test GPT-5.1 and various open-source LLMs (up to 30B parameters) on a consumer-grade GPU to estimate the gap between open-source agentic methodologies and leading human-in-the-loop autoformalisation pipelines. The prevailing open-source agent (using GPT-OSS-20B) obtained a compilation rate $\approx 1/3$ that of the FormalScience approach.

We characterise semantic formalisation errors by defining alignment drift categories, explore errors qualitatively, then determine what verification guarantees can be made depending on the specific drift type supported by domain-specific quantitative analysis. Despite the perfect formal verification rate, notational collapse occurs in most QM proofs, which instead guarantees the verification of surrogate solutions involving simple mathematical objects. Abstraction elevation guarantees meaningful physics computation is circumvented with oversimplified proofs of goal formulae, or is abstracted away with general algebraic proofs which does not necessarily respect the scientific context. We also find drift can be beneficial. In the case of implicit premise selection, without any other drift types, proofs are enriched by supporting premises and rigorous argumentation. Overall, we believe our work will support the development of LLM-based alignment metrics, guide the construction of fully automated formalisation agents, and accelerate scientific formalisation. While the FormalScience pipeline is domain-agnostic, the empirical analysis presented here is restricted to physics; applying FormalScience to other scientific domains (\textit{e.g.} chemistry, biology) remains future work.

\section*{Limitations}

\textbf{Dataset scale and scope.} FormalPhysics comprises 200 examples focused on quantum mechanics and electromagnetism at university level. This scope, while sufficient for benchmark evaluation, limits generalisability to other physics subdomains (e.g., statistical mechanics, general relativity) and other sciences.\newline

\noindent \textbf{Formal system constraints.} As analysed in Section 5, Lean4's Mathlib lacks native support for vector calculus and Dirac notation, necessitating semantic drift in formalisations. The formal proofs therefore verify algebraic consistency within abstract structures rather than the complete physical derivations. Addressing this limitation requires either extending Mathlib's physics coverage or developing physics-specific formal libraries.\newline

\noindent\textbf{Resource requirements.} The FormalScience pipeline required approximately one month of expert effort. The agentic baselines required over 100 hours of GPU compute per 200 examples on consumer hardware. These costs may limit broader adoption and scaling to larger corpora. \newline

\noindent\textbf{Evaluation methodology.} Alignment metrics rely on LLM-as-a-judge evaluation, which may not capture all dimensions of semantic preservation. An inter-judge robustness analysis with an independent 7B judge (Appendix~\ref{app:judge-robustness}) confirms our central findings but reveals that some secondary claims (\textit{e.g.} score invariance under self-refinement, emergence effect size) are judge-dependent. The drift categorisation taxonomy we propose is one principled decomposition but not necessarily complete or unique.\newline

\noindent\textbf{Temporal validity.} Results reflect specific LLM versions and Lean4/Mathlib configurations. Model capabilities and library coverage evolve rapidly, and our findings should be interpreted in this context.

\bibliography{latex/custom}

\appendix

\section{Implementation of FormalScience pipeline to construct FormalPhysics}
\label{app:autoformalisation-pipeline-details}

As described in Alg.~\ref{alg:autoformalisation}, the FormalScience pipeline requires in-context examples of informal statements and proofs, and a larger collection of informal proofs we aim to formalise, generate informal statements for, and extend or otherwise alter (\textit{e.g.} add annotations) guided by the few-shot examples. 

\paragraph{Data preparation.} In the case of the physics examples, we curate 5 such gold-standard statement-proof pairs (see Fig.~\ref{fig:few-shot}) and randomly select 200 examples from a related dataset of derivations~\cite{meadows-etal-2024-exploring}. We randomise the 200 examples and group them into batches of 5. Each batch, and the few-shot examples, are now in the required format for input to FormalScience. This ultimately results in 40 few-shot prompts (generated automatically with a Python script) for the initial generation of expanded statements and informal answers (see Section~\ref{sec:formalscience}) using an LLM.

\paragraph{Stage 1: Generating informal statements and (expanded) proofs.} We used GPT-5.1 in thinking mode to generate an intermediate dataset comprising of an informal "question" (statement) and "answer" (proof) from the initial few-shot prompt template (Tab.~\ref{tab:formalscience-prompts}). In particular, the answers were expanded to include NL step annotations in a significant number of cases. After each few-shot prompt, the human expert evaluated the alignment between statement and proof, or otherwise prompted GPT to improve the alignment. Each of the 40 resulting raw LLM outputs contained alternating questions (Q1 - Q5) and answers (A1 - A5). These were split up via a post-processing script to form an intermediate dataset of 40 examples where each example contains 5 dictionaries per derivation. Each dictionary contains the "field" (\textit{e.g.} electromagnetism), the question, and the answer.

\paragraph{Stage 2: Generating formal proofs in Lean4.} Each example from the intermediate dataset from the previous stage was input to a formalisation prompt template (see Tab~\ref{tab:formalscience-prompts}).

Our implementation diverges at this point into two complimentary approaches, as we used the ChatGPT interface to produce around 1/3 of examples, and Claude Code (through VSCode) to generate the rest. 

The ChatGPT approach required manually copying the outputted formal code, compiling it in Lean, then prompting GPT (within the same session) with the raw compilation errors. We removed non-fatal warnings to reduce context window limitations. Per initial formalisation prompt, this approach required approximately 1-2 hours and several rounds of prompts to generate outputs without compilation errors. Occasionally GPT had to be reminded that incorrect imports did not mean the Lean environment was incomplete. 

The Claude Code approach (recommended) included less manual compilation and human intervention. A custom Lean compilation (Python) script was written (\textit{i.e.} $\mathcal{L}(C)$, Section~\ref{sec:formalscience}), and instructions on how to use it, and handle its output, were appended to the end of the formalisation prompt template. Importantly, due to the context window compactification implemented frequently during this approach, we also saved the exact formalisation prompt in a separate text file in the same folder as the Lean compilation tool, which we found aided the later alignment stage. 

Each generated formal code output is prefixed with ``C1-C5'', due to the formalisation prompt template containing questions ``Q1-Q5'' and informal solutions ``A1-A5''.

\paragraph{Stage 3: Iterative alignment.}

Upon successful compilation without errors, our approach to iterative alignment was centered around a single fixed prompt: 

\textit{"How well do C1-C5 align with A1-A5, the Requirements, and the Acceptance criteria?"}

In either the Claude Code or ChatGPT interface approaches, the human expert evaluates whether the resulting alignment analysis is acceptable. Reaching this point in the pipeline is in itself an iterative process we call \textit{patience}, $\mathcal{P}$, in Alg.~\ref{alg:autoformalisation}. We used a maximum patience $\mathcal{P} = 3$ before considering the formal code ``well-aligned''. 

If the formal code was not aligned, an additional fixed prompt (\textit{``Make the suggested improvements and ensure C1-C5 aligns with A1-A5, the Requirements, and the Acceptance criteria.''}) was used in either case. Claude Code had to additionally be instructed to iteratively use the Lean compilation tool, while this was manually achieved by the human expert for ChatGPT. This essentially restarts Stage 2 and ticks patience $\mathcal{P}$ . Notably, for GPT, we found only compiling the poorly-aligned code vastly accelerated this process. 

\paragraph{Stage 4: Post-processing and formal re-verification.} Both approaches converge to the same post-processing method. Each formal code output from Stage 3 (a total of 40) contains 5 separate formal code proofs C1-C5, where Mathlib imports for all proofs are combined at the top of the file, within the same block. We use a Python script to separate out all proofs into 5 separate files with identical import blocks. We use another script to unify all questions, informal answers, formal proofs, and physics subdomain categories into a dataset of 200 examples.

We compile each formal proof in this dataset to determine if the separation script introduced any errors, making a list of example IDs with new errors. We iteratively improve each proof, beginning from Stage 3 in each case, but reword the alignment prompts to consider only one input question (Q), informal answer (A), and formal code (C) at a time. The resulting examples are added back into the dataset, which finalises our implementation of the FormalScience pipeline used to generate the FormalPhysics dataset.

\begin{algorithm*}[h!]
\caption{FormalScience}
\label{alg:autoformalisation}
\begin{algorithmic}[1]
\Require Derivation batches $\mathcal{D}$, batch size $B$ (with $B' \in [1, B]$), gold-standard examples $\mathcal{D}^* = [(Q_i, A_i)]_{i=1}^{N'}$, patience $\mathcal{P}$. Here, questions $Q$ and informal answers $A$ are respectively equivalent to informal statements $S$ and informal proofs $P$ in the main text. 
\Ensure Formally verified corpus $Z = \{(Q, A, C) : \mathcal{L}(C) = (0, \varepsilon)\}$

\Statex \textit{\textbf{Stage 1: Informal QA Generation and Alignment}}
\State $X \gets []$
\For{each batch $d \in \mathcal{D}$}
    \State $x \gets S\big(\mathcal{M}(T_{fs}(d, \mathcal{D}^*); P_a)\big)$ \Comment{Few-shot QA generation with human alignment}
    \State $X \gets X \cup \{x\}$
\EndFor

\Statex
\Statex \textit{\textbf{Stage 2: Code Generation and Iterative Correction}}
\State $Y \gets []$
\For{each $x = [(Q_i, A_i)]_{i=1}^{B'} \in X$}
    \State $C^{(0)} \gets \mathcal{M}'(T_g(x))$ \Comment{Initial code generation}
    \State $t \gets 0$
    \While{$\mathcal{L}(C^{(t)}) = (1, e)$} \Comment{Compilation fails}
        \State $C^{(t+1)} \gets \mathcal{M}'(T_c(x, C^{(t)}, e))$ \Comment{Error-guided correction}
        \State $t \gets t + 1$
    \EndWhile
    \State $C \gets C^{(t)}$ \Comment{$C = \mathcal{R}(C^{(0)})$}
    
    \Statex
    \Statex \textit{\textbf{Stage 3: Formal Language Alignment}}
    \State $k \gets 0$
    \While{$\mathcal{H}^{(k)} = 1$ \textbf{and} $k < \mathcal{P}$} \Comment{Human rejects alignment}
        \State $C' \gets \mathcal{M}'(\mathcal{M}'(T_g(x), C^{(k)}), P_g)$ \Comment{Alignment improvement}
        \State $C^{(k+1)} \gets \mathcal{R}(C')$ \Comment{Re-verify compilation}
        \State $k \gets k + 1$
    \EndWhile
    \State $y \gets S'(x, C^{(k)})$ \Comment{Split into $(Q_i, A_i, C_i)$ tuples}
    \State $Y \gets Y \cup \{y\}$
\EndFor

\Statex
\Statex \textit{\textbf{Stage 4: Post-processing and Final Verification}}
\State $Z \gets \text{flatten}(Y)$
\For{each $(Q, A, C) \in Z$}
    \If{$\mathcal{L}(C) = (1, e)$} \Comment{Post-processing introduced errors}
        \State $(Q, A, C) \gets$ apply Eq.~\ref{eq:alignment_iteration} \Comment{Re-align and correct}
    \EndIf
\EndFor
\State \Return $Z$
\end{algorithmic}
\end{algorithm*}

\section{Prompts}
\label{app:prompts}
\begin{figure}
    \centering
    \includegraphics[width=1\linewidth]{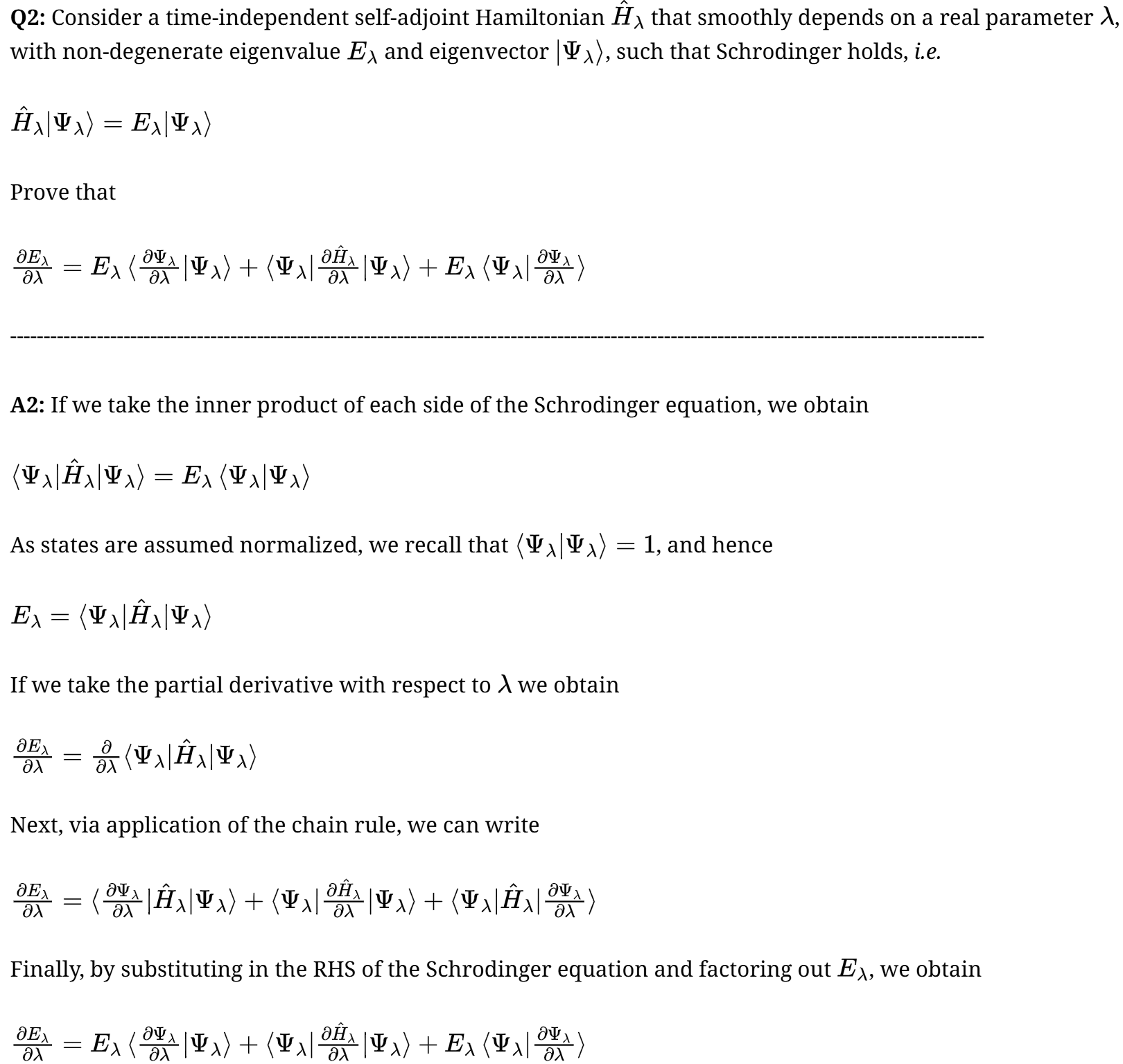}
    \caption{A (human-written) in-context example used in a few-shot prompt to automatically provide physical context to other examples with a similar degree of depth.}
    \label{fig:few-shot}
\end{figure}

We provide the few-shot template, the prompt for FormalScience, and prompt for testing LLMs in Figure~\ref{fig:few-shot}, Table~\ref{tab:formalscience-prompts}, Table~\ref{tab:zero-shot-prompts-and-refinement}, respectively.

\begin{table*}[h!]
\centering
\small
\begin{tabularx}{\textwidth}{>{\raggedright\arraybackslash}p{2.2cm} >{\raggedright\arraybackslash}p{2.5cm} >{\raggedright\arraybackslash}X >{\raggedright\arraybackslash}p{2.8cm} >{\raggedright\arraybackslash}p{2.5cm}}
\toprule
\textbf{Prompt Name} & \textbf{Purpose} & \textbf{Template Structure} & \textbf{Input Variables} & \textbf{Output Expected} \\
\midrule
Informal Expansion Prompt & 
Convert equation-only derivations into contextually-rich Q\&A pairs & 
\textit{``The following 5 questions (Q1--Q5) and respective answers (A1--A5) are few-shot examples...''} followed by 5 quantum Q\&A pairs, then \textit{``Now, the following \textbf{equation-only} derivations (D6--D10) represent the underlying equational reasoning of a Physics derivation. You must convert each derivation into a \textbf{physically-correct} and \textbf{contextually-enriched} Question (Q6--Q10) and Answer (A6--A10) pair...''} followed by 5 derivations and closing instruction to ensure one equality per equation with correct physical meaning and standard notation. & 
\begin{itemize}[nosep,leftmargin=*]
    \item 5 few-shot Q\&A pairs.
    \item 5 equation-chain derivations (\textit{i.e.} a batch).
\end{itemize} & 
5 new Q\&A pairs (Q6--Q10, A6--A10) with physics context and properly formatted \LaTeX{} equations \\
\midrule
Lean Formalization Prompt & 
Autoformalize informal physics derivations into compilable Lean~4 + Mathlib proofs & 
\textbf{Header:} Task description requesting compilable Lean~4 proofs without axioms. \textbf{Q/A Block:} 5 pairs formatted as \textit{``\textbf{Q$i$:} \{question\} \textbf{A$i$:} \{answer\}''}. \textbf{Requirements:} (1)~No \texttt{axiom}/\texttt{sorry}; (2)~Use Mathlib theorems; (3)~Explicit physics modelling; (4)~Single compilable file; (5)~Include docstrings and clear theorem names; (6)~Prefer \texttt{calc} blocks over \texttt{simp}; (7)~Deterministic rewrites. \textbf{Deliverables:} One file with C1--C5. \textbf{Acceptance:} Must compile with no axioms. & 
\begin{itemize}[nosep,leftmargin=*]
    \item 5 expanded informal Q\&A pairs from previous LLM output. 
\end{itemize} & 
Single Lean~4 file containing 5 theorems (C1--C5) with Mathlib imports, docstrings, explicit hypotheses, and complete proofs \\
\bottomrule
\end{tabularx}
\caption{Prompt Templates Used in the FormalScience Pipeline}
\label{tab:formalscience-prompts}
\end{table*}

\begin{table*}[h!]
    \centering
    \small
    \begin{tabular}{p{0.1\textwidth} p{0.8\textwidth}}
        \toprule
        Task & Content\\
        \midrule
        Zero-Shot Statement Autoformalisation & 
        You are an expert in formal language Lean4.\newline
        You will be given a physics statement and its proof written in natural language and LaTeX symbols.\newline
        Your task is to provide the formal code of the given natural language physics statement and its proof in Lean4 with the following instructions:\newline
        1. You should give the formal code directly without any additional comments or explanations. If the given physics statement is a theorem or lemma, omit the formal proof and use the default 'sorry' mode in the formal code.\newline
        2. In case that you need to import any necessary preambles, you should not import any fake (non-exist) preambles.\newline
        3. You should wrap the formal code in a way illustrated as the following:\newline
        \%\%\%\%\%\%\%\%\%\%\newline
        Your Formal Code\newline
        \%\%\%\%\%\%\%\%\%\%\newline
        Strictly follow the instructions that have been claimed.\newline
        Natural language statement: \{nl\_statement\}\newline
        Give me the Lean4 formal code of the statement:\\
        \midrule
        Self-Refinement with Error Feedback & 
        You are an expert in formal language Lean4.\newline
        You will be given a physics statement and its proof written in natural language and LaTeX symbols.\newline
        You will also be given a formal code which attempted to describe the given physics statement in Lean4.\newline
        Your task is to refine the given formal code to make it correct while maintaining the alignment with the given natural language physics statement.\newline
        Here are some instructions for your task:\newline
        1. You should give the formal code directly without any additional comments or explanations. If the given physics statement is a theorem or lemma, omit the formal proof and use the default 'sorry' mode in the formal code.\newline
        2. In case that you need to import any necessary preambles, you should not import any fake (non-exist) preambles.\newline
        3. You should wrap the formal code in a way illustrated as the following:\newline
        \%\%\%\%\%\%\%\%\%\%\newline
        Your Formal Code\newline
        \%\%\%\%\%\%\%\%\%\%\newline
        Strictly follow the instructions that have been claimed.\newline
        Natural language statement: \{nl\_statement\}\newline
        There are some Lean4 formal codes describing the given physics statement: \{formal\}\newline
        You should refine the formal code for your task to make it correct.\newline
        Here are some feedbacks about the formal code which can be used to help your task: \{According to the theorem prover, the error details of the provided formal code are:\newline 
        error\_details\newline
        \}\\
        \bottomrule
    \end{tabular}
    \caption{Prompts.}
    \label{tab:zero-shot-prompts-and-refinement}
\end{table*}

\section{Derivation of the agentic code generation pipeline}
\label{app:agent-derivation}

\begin{algorithm*}[h!]
\caption{Agentic Code Generation Pipeline}
\label{alg:agentic}
\footnotesize
\begin{algorithmic}[1]
\Require Physics question $x$, initial generation template $T_g$, regeneration template $T_r$, patch template $T_p$, max initial attempts $N = 25$, max correction steps $N_{\max} = 25$, forbidden tokens $F = \{\partial, \nabla, \dot{x}, \dot{y}, \dot{z}, \dagger, \backslash\backslash, \texttt{`}\}$
\Ensure Compilable code $C$ such that $\mathcal{L}(C) = (0, \varepsilon)$, or failure

\Statex
\Statex \textbf{Stage 1: Initial Generation with Surface Guard}
\State $i \gets 0$
\While{$i < N$}
    \State $C^{(i)} \gets \text{extract}\bigl(\mathcal{M}(T_g(x))\bigr)$ \Comment{Eq. 9: Generate and extract code}
    \State $(g, r^{(i)}) \gets G(C^{(i)})$ \Comment{Surface guard validation}
    \If{$g = 1$} \Comment{Guard passed}
        \State \textbf{goto} Stage 2
    \Else
        \State $C^{(i+1)} \gets \text{extract}\bigl(\mathcal{M}(T_r(x, r^{(i)}, \bot))\bigr)$ \Comment{Eq. 10: Regenerate with feedback}
    \EndIf
    \State $i \gets i + 1$
\EndWhile
\State \Return \textsc{Failure} \Comment{No valid candidate in $N$ attempts}

\Statex
\Statex \textbf{Stage 2: Iterative Compilation and Error Correction}
\State $C \gets C^{(i^*)}$ where $i^* = \min\{i : G(C^{(i)}) = (1, \varepsilon)\}$
\State $t \gets 0$
\While{$t < N_{\max}$}
    \State $(s, e) \gets \mathcal{L}(C^{(t)})$ \Comment{Eq. 5: Lean compilation}
    \If{$s = 0$} \Comment{Compilation succeeded}
        \State \Return $C^{(t)}$
    \EndIf
    \State $\kappa(e) \gets$ \Call{Categorize}{$e$} \Comment{Error categorisation}
    
    \Statex
    \If{$\kappa(e) \in E_{\text{struct}}$} \Comment{Structural errors: syntax, unknown\_id, missing\_module}
        \State $C' \gets \text{extract}\bigl(\mathcal{M}(T_r(x, \kappa(e), e))\bigr)$ \Comment{Eq. 11: Full regeneration}
        \If{$G(C') = (1, \varepsilon)$} \Comment{Guard passes}
            \State $C^{(t+1)} \gets C'$
        \Else
            \State Log regeneration failure; $C^{(t+1)} \gets C^{(t)}$
        \EndIf
    \Else \Comment{Semantic errors: type\_mismatch, unsolved\_goals, other}
        \State $C^{(t+1)} \gets \mathcal{A}_{\text{patch}}\bigl(T_p(\text{numberlines}(C^{(t)}), e)\bigr)$ \Comment{Eq. 12: Patch agent}
    \EndIf
    \State $t \gets t + 1$
\EndWhile
\State \Return $C^{(N_{\max})}$ \Comment{Return best effort after max steps}

\Statex
\Statex \textbf{-- Helper Definitions --}
\Statex
\Function{Categorize}{$e$}
    \State $E_{\text{struct}} \gets \{\texttt{syntax}, \texttt{unknown\_id}, \texttt{missing\_module}\}$
    \State $E_{\text{sem}} \gets \{\texttt{type\_mismatch}, \texttt{unsolved\_goals}, \texttt{other}\}$
    \State \Return category $\kappa(e) \in E_{\text{struct}} \cup E_{\text{sem}}$ via pattern matching
\EndFunction

\Statex
\Function{$G$}{$C$} \Comment{Surface guard: $G : \text{Code} \to \{0,1\} \times \Sigma^*$}
    \If{$\exists \, f \in F : f \in C$} \Return $(0, \text{``forbidden token } f \text{''})$ \EndIf
    \If{$\texttt{sorry} \in C \lor \texttt{axiom} \in C$} \Return $(0, \text{``incomplete proof''})$ \EndIf
    \If{$|C|_{\texttt{/-}} \neq |C|_{\texttt{-/}}$} \Return $(0, \text{``unmatched delimiters''})$ \EndIf
    \If{imports not correctly ordered} \Return $(0, \text{``import ordering''})$ \EndIf
    \State \Return $(1, \varepsilon)$
\EndFunction

\end{algorithmic}
\end{algorithm*}

Let $T_g$ be the initial prompt template for physics question $x$. The initial generation phase attempts up to $N = 25$ iterations:
\begin{equation}
    C^{(0)} = \texttt{extract}\Big(\mathcal{M}\big(T_g(x)\big)\Big)
\end{equation}
where $\texttt{extract}: \Sigma^* \to \text{Code}$ removes markdown fences from the LLM output. Each candidate $C^{(i)}$ is passed through a surface guard $\mathcal{G}: \text{Code} \to \{0,1\} \times \Sigma^*$ which performs syntax validation before compiling in Lean. The guard returns $(1, \varepsilon)$ if the code passes all heuristic checks, or $(0, r)$ with rejection reason $r$ otherwise. These checks enforce constraints already specified in $T_g$ but which the LLM may violate: absence of forbidden tokens $\mathcal{F} = \{\partial, \nabla, \dot{x}, \dot{y}, \dot{z}, \dagger, \texttt{\textbackslash\textbackslash}, \texttt{`}\}$, absence of incomplete proof markers (\texttt{sorry}, \texttt{axiom}), balanced comment delimiters, and correct import ordering. If the guard fails, the agent regenerates with feedback:
\begin{equation}
    C^{(i+1)} = \texttt{extract}\Big(\mathcal{M}\big(T_r(x, r^{(i)}, \bot)\big)\Big)
\end{equation}
where $\mathcal{G}(C^{(i)}) = (0, r^{(i)})$, the $T_r$ is a regeneration template that appends the rejection reason to the base prompt, and $\bot$ denotes the absence of a compiler error (since compilation has not yet been attempted). The phase terminates at $i^* = \min\{i : \mathcal{G}(C^{(i)}) = (1, \varepsilon)\}$ or fails if no valid candidate is found within $N$ attempts. \newline

\noindent Once initial generation succeeds, the agent enters an iterative compilation loop for up to $N_{\max} = 25$ steps. At each step $n$, the Lean compiler $\mathcal{L}(C^{(n)})$ returns $(0, \varepsilon)$ on successful compilation or $(1, e)$ with error message $e$ otherwise (\textit{i.e.} Eq.~\ref{eq:compile_lean}). When compilation fails, a categorisation function $\kappa: \Sigma^* \to \mathcal{E}$ maps the error to one of six categories via pattern matching $\mathcal{E} = \{\texttt{syntax}, \texttt{unknown\_id}, \texttt{missing\_module},\\ \texttt{type\_mismatch}, \texttt{unsolved\_goals}, \texttt{other}\}$. These categories are split into structural errors $\mathcal{E}_{\text{struct}} = \{\texttt{syntax}, \texttt{unknown\_id}, \texttt{missing\_module}\}$ and semantic errors $\mathcal{E}_{\text{sem}} = \{\texttt{type\_mismatch}, \texttt{unsolved\_goals}, \texttt{other}\}$, which determine the repair strategy. \newline

\noindent For structural errors, the agent performs full regeneration using the primary LLM:
\begin{equation}
    C^{(t+1)} = \texttt{extract}\Big(\mathcal{M}\big(T_r(x, \kappa(e), e)\big)\Big)
\end{equation}
if $\kappa(e) \in \mathcal{E}_{\text{struct}}$ where $T_r$ now includes both the error category and the compiler message, instructing the model to rewrite the entire file. The regenerated code must again pass the surface guard; if $\mathcal{G}(C^{(t+1)}) = (0, \_)$, the regeneration is discarded and the step is logged as failed. \newline

\noindent For semantic errors, the agent uses a specialised patch agent $\mathcal{A}_{\text{patch}}$ equipped with a unified diff tool:
\begin{equation}
    C^{(t+1)} = \mathcal{A}_{\text{patch}}\Big(T_p\big(\texttt{number}(C^{(t)}), e\big)\Big)
\end{equation}
if $\kappa(e) \in \mathcal{E}_{\text{sem}}$ where $\texttt{number}: \text{Code} \to \text{Code}$ prepends line numbers and $T_p$ is a patch template requesting a minimal unified diff. The patch agent operates in a ReAct loop, generating a diff and applying it via the \texttt{apply\_unified\_diff} tool. This reflects the intuition that structural errors indicate fundamental misunderstanding requiring complete regeneration, while semantic errors (type mismatches, unsolved goals) are often addressable through localised edits to tactics or expressions. \newline

\noindent The correction loop terminates when $\mathcal{L}(C^{(t)}) = (0, \varepsilon)$ or when $n = N_{\max}$. Unlike the human-in-the-loop pipeline (Alg.\ref{alg:autoformalisation}, Stage 3), this approach does not perform formal language alignment. The agentic code generation pipeline used to obtain the relevant baselines is described in Alg.~\ref{alg:agentic}.

\section{Supplementary Analysis}

\subsection{LLM-Judge Robustness Analysis}
\label{app:judge-robustness}

To assess the robustness of our LLM-as-a-judge alignment evaluation, we conduct an inter-annotator agreement analysis using a second, independent judge (Qwen2.5-Coder-7B-Instruct) alongside our primary judge (GPT-4.1-mini). Both judges evaluated all 200 items $\times$ 5 models $\times$ 2 settings (zero-shot and self-refinement) $\times$ 3 alignment metrics (FQ, LP, MC), yielding ${\sim}6{,}000$ paired binary judgments. Full 7B judge results are reported in Table~\ref{tab:llm-baselines-7b}.

\paragraph{Judges detect the same underlying quality signal despite different calibration.} Pooling all (item $\times$ model) pairs per setting and metric gives 1,000 paired binary observations per condition. The phi coefficient is positive and highly significant across all six conditions ($\phi \in [0.28, 0.37]$, all $p < 10^{-19}$), confirming both judges respond to the same underlying quality signal.

\begin{table}[h!]
\centering
\small
\begin{tabular}{l c c c}
\toprule
\textbf{Setting} & \textbf{FQ} & \textbf{LP} & \textbf{MC} \\
\midrule
Zero-Shot & 0.28 & 0.33 & 0.37 \\
Self-Refinement & 0.30 & 0.32 & 0.35 \\
\bottomrule
\end{tabular}
\caption{Phi coefficients between GPT-4.1-mini and Qwen2.5-Coder-7B-Instruct judges ($n{=}1{,}000$ per cell, all $p < 10^{-19}$).}
\label{tab:phi-coefficients}
\end{table}

\paragraph{Disagreement is structured and asymmetric.} The 7B judge is systematically more conservative. For the two strongest models (GPT-5.1 and GPT-OSS-20B), over 95\% of inter-judge disagreements take the form GPT{=}True / 7B{=}False. When the conservative 7B judge accepts an item, the GPT judge almost always agrees (91--100\% for GPT-5.1, 89--95\% for GPT-OSS-20B). The 7B-positive items therefore form a high-confidence consensus subset.

\paragraph{Model-level rankings are partially preserved.} Kendall's $\tau$ across all six (metric $\times$ setting) comparisons ranges from 0.2 to 1.0 (median 0.80). Five of six comparisons yield $\tau \geq 0.6$. Both judges unanimously rank GPT-5.1 first on every metric in every setting. Rank instability is confined to the middle and bottom of the ranking, where the 7B judge's compressed score distributions (four models within a 5pp band) make fine-grained distinctions unreliable.

\paragraph{Implications for paper claims.}
The alignment-validity trade-off (Section~\ref{sec:formalscience}) is \textit{supported}: under the 7B judge, the Spearman correlation between FV and mean alignment remains near zero ($\rho \in [-0.10, 0.30]$, all $p > 0.6$). The claim that scores are ``effectively unchanged with self-refinement'' is \textit{judge-dependent}: it holds for the GPT judge but not universally (GPT-5.1 gains +9.0pp under the 7B judge). The emergence effect size is \textit{directionally preserved but magnitude-reduced}: the ratio of GPT-5.1 to DeepSeek-Prover-7B mean alignment is $3.1\times$ under the GPT judge but $2.2\times$ under the 7B judge (zero-shot). Kimina-7B's compilation shortcut exploitation is \textit{supported} under both judges.

\begin{table*}[h!]
\centering
\small
\caption{Detailed semantic drift analysis: Physics notation versus Lean4 formalisation. This table provides extended examples showing how mathematical objects transform during autoformalisation.}
\label{tab:drift_detailed}
\renewcommand{\arraystretch}{1}
\begin{tabular}{@{}p{0.6cm}p{6.8cm}p{7.2cm}@{}}
\toprule
\textbf{ID} & \textbf{Physics (Informal)} & \textbf{Lean4 (Formal)} \\
\midrule
\multicolumn{3}{l}{\textbf{Pattern A: Quantum Operators $\to$ Scalars/Algebra Elements}} \\
\midrule

4 &
\textbf{Statement:} $N = a^\dagger a = \frac{1}{2}(q-ip)(q+ip)$

\textbf{Key objects:}
\begin{itemize}[nosep,leftmargin=*]
\item $q, p$: Position/momentum operators with $[q,p] = i$
\item $a, a^\dagger$: Ladder operators (non-commuting)
\item $N$: Number operator
\end{itemize}

\textbf{Physics content:} Operator algebra on Hilbert space &

\begin{lstlisting}[basicstyle=\ttfamily\tiny,breaklines=true]
theorem C5_number_operator_expand
    (q p : C) :  -- Complex numbers!
    let s := (1 : R) / Real.sqrt 2
    let a    := s * (q + I * p)
    let adag := s * (q - I * p)
    adag * a = s*s * (p^2 - I*(p*q) 
                    + I*(q*p) + q^2)
\end{lstlisting}

\textbf{Drift:} $[q,p]=i$ \emph{not enforced}; proof holds for any complex numbers \\
\midrule

14 &
\textbf{Statement:} Heisenberg uncertainty $\Delta X \cdot \Delta P \geq \hbar/2$

\textbf{Key objects:}
\begin{itemize}[nosep,leftmargin=*]
\item $|f\rangle, |g\rangle$: Fluctuation kets in Hilbert space
\item $[\hat{X}, \hat{P}] = i\hbar$: Canonical commutation relation
\item Cauchy-Schwarz on inner products
\end{itemize}

\textbf{Physics content:} Fundamental quantum bound &

\begin{lstlisting}[basicstyle=\ttfamily\tiny,breaklines=true]
theorem C1
    {E : Type*} [InnerProductSpace C E]
    {f g : E} {hbar : R}
    (hbar_nonneg : 0 <= hbar)
    (hcomm : inner f g - inner g f 
           = I * (hbar : C)) :
    ||f|| * ||g|| >= hbar / 2
\end{lstlisting}

\textbf{Drift:} Commutator relation is a \emph{hypothesis}, not derived from $[\hat{X},\hat{P}]$ \\
\midrule

12 &
\textbf{Statement:} $\frac{d\hat{x}(t)}{dt} = \frac{i}{\hbar}[\hat{H}, \hat{x}(t)]$

\textbf{Key objects:}
\begin{itemize}[nosep,leftmargin=*]
\item $e^{i\hat{H}t/\hbar}$: Unitary time evolution operator
\item $\hat{x}(t) = U(t)\hat{x}U^\dagger(t)$: Heisenberg picture
\item Product rule on operator exponentials
\end{itemize}

\textbf{Physics content:} Quantum dynamics &

\begin{lstlisting}[basicstyle=\ttfamily\tiny,breaklines=true]
theorem C3
    {A : Type*} [NormedAlgebra R A]
    (c : R) {H x0 : A}
    (U W : R -> A)
    (hU : forall t, deriv U t = c*(H*U t))
    (hW : forall t, deriv W t = -c*(W t*H))
    (hUd : forall t, DifferentiableAt R U t)
    (hWd : forall t, DifferentiableAt R W t) :
    forall t, deriv (s => U s * x0 * W s) t
        = c * (H*(U t*x0*W t) - (U t*x0*W t)*H)
\end{lstlisting}

\textbf{Drift:} Exponential structure assumed via \texttt{hU}, \texttt{hW}; works in any normed algebra \\
\midrule
\multicolumn{3}{l}{\textbf{Pattern B: Vector Calculus $\to$ Abstract Linear Maps}} \\
\midrule

5 &
\textbf{Statement:} $\hat{\mathbf{r}} \cdot \nabla f = \frac{\partial f}{\partial r}$

\textbf{Key objects:}
\begin{itemize}[nosep,leftmargin=*]
\item $\nabla f = \frac{\partial f}{\partial r}\hat{\mathbf{r}} + \frac{1}{r}\frac{\partial f}{\partial\theta}\hat{\boldsymbol{\theta}} + \frac{1}{r\sin\theta}\frac{\partial f}{\partial\varphi}\hat{\boldsymbol{\varphi}}$
\item Spherical coordinate basis $\{\hat{\mathbf{r}}, \hat{\boldsymbol{\theta}}, \hat{\boldsymbol{\varphi}}\}$
\end{itemize}

\textbf{Physics content:} Coordinate geometry &

\begin{lstlisting}[basicstyle=\ttfamily\tiny,breaklines=true]
theorem C1_radial_projection
  {E : Type*} [InnerProductSpace R E]
  {er e_th e_ph : E} {ar ath aph : R}
  (hnorm : ||er|| = 1)
  (horth_r_th : inner er e_th = 0)
  (horth_r_ph : inner er e_ph = 0) :
  inner er (ar*er + ath*e_th + aph*e_ph) = ar
\end{lstlisting}

\textbf{Drift:} Gradient formula becomes hypothesis; spherical coords $\to$ abstract orthonormal basis \\
\midrule

24 &
\textbf{Statement:} Gauss's law from $\nabla^2\phi = 4\pi G\rho$

\textbf{Key objects:}
\begin{itemize}[nosep,leftmargin=*]
\item $\int_V \nabla^2\phi\,dV = 4\pi G \int_V \rho\,dV$
\item Divergence theorem: $\int_V \nabla\cdot\mathbf{F}\,dV = \oint_{\partial V} \mathbf{F}\cdot d\mathbf{S}$
\end{itemize}

\textbf{Physics content:} Field theory &

\begin{lstlisting}[basicstyle=\ttfamily\tiny,breaklines=true]
theorem C5_gauss_from_poisson
    (G : R) (S V1 V2 : R)
    (hPoisson : V1 = (4 * pi * G) * V2)
    (hDivThm : S = V1) :
    S = (4 * pi * G) * V2 := by
  simpa [hPoisson] using hDivThm
\end{lstlisting}

\textbf{Drift:} All integration/divergence content in hypotheses; proof is trivial substitution \\
\midrule
\multicolumn{3}{l}{\textbf{Pattern C: Integration $\to$ Limit/Algebraic Statement}} \\
\midrule

10 &
\textbf{Statement:} $W = \int_{r_0}^{\infty} \frac{GMm}{r^2}\,dr = \frac{GMm}{r_0}$

\textbf{Key objects:}
\begin{itemize}[nosep,leftmargin=*]
\item Definite integral of inverse-square force
\item Improper integral at infinity
\end{itemize}

\textbf{Physics content:} Work-energy theorem &

\begin{lstlisting}[basicstyle=\ttfamily\tiny,breaklines=true]
theorem C1
    (G M m r0 : R) (hr0 : 0 < r0) :
  Tendsto (fun R => (G*M*m)*(r0^-1 - R^-1)) 
          atTop (nhds ((G*M*m) / r0)) := by
  have h_inv : Tendsto (R^-1) atTop (nhds 0) 
             := tendsto_inv_atTop_zero
  ...
\end{lstlisting}

\textbf{Drift:} Integral $\to$ limit statement; uses \texttt{tendsto\_inv\_atTop\_zero} lemma \\
\midrule

27 &
\textbf{Statement:} $U = q\phi(r)$ from $W = \int_\infty^r \mathbf{F}\cdot d\mathbf{l}$

\textbf{Key objects:}
\begin{itemize}[nosep,leftmargin=*]
\item Line integral of electric force
\item Reference $\phi(\infty) = 0$
\end{itemize}

\textbf{Physics content:} Potential energy definition &

\begin{lstlisting}[basicstyle=\ttfamily\tiny,breaklines=true]
theorem C3_potential_energy_from_reference
    (q r : R) (phi : R -> R) :
    q * (phi r - 0) = q * phi r := by
  simpa [sub_zero]
\end{lstlisting}

\textbf{Drift:} Entire integration content abstracted; proof is \texttt{sub\_zero} \\
\bottomrule
\end{tabular}
\label{tab:detailed-alignment-table}
\end{table*}

\end{document}